\documentclass[10pt,twocolumn,letterpaper]{article}

\usepackage{cvpr}
\usepackage{eso-pic}
\usepackage{times}
\usepackage{epsfig}
\usepackage{graphicx}
\usepackage{amsmath}
\usepackage{amssymb}
\usepackage{fancyhdr}

\usepackage[breaklinks=true,bookmarks=false]{hyperref}
\usepackage{cite}

\cvprfinalcopy 

\usepackage{lipsum}

\newcommand\blfootnote[1]{%
  \begingroup
  \renewcommand\thefootnote{}\footnote{#1}%
  \addtocounter{footnote}{-1}%
  \endgroup
}

\begin{document}
\title{Using Artificial Tokens to Control Languages for Multilingual Image Caption Generation}

\author{Satoshi Tsutsui \hspace{1in} David Crandall\\
School of Informatics and Computing\\
Indiana University\\
{\tt\small \{stsutsui,djcran\}@indiana.edu}
}

\maketitle

\begin{abstract}
Recent work in computer vision has yielded impressive results in
automatically describing images with natural language.
   Most of these systems generate captions in a
   single language, requiring multiple language-specific models to build
   a multilingual captioning system. We propose a very simple technique
   to build a single
   unified model across languages, using artificial tokens to control
   the language, making the captioning system more compact. We
   evaluate our approach on generating English and Japanese captions,
   and show that a typical neural captioning architecture is capable
   of learning a single model that can switch between two different
   languages.  \blfootnote{This work appears as an Extended Abstract at the 2017 CVPR Language and Vision Workshop.}
\end{abstract}



\section{Introduction}

A key problem in the intersection between computer vision and natural
language processing is to automatically describe an image in natural
language, and recent work has shown exciting progress using deep
neural network-based models~\cite{vinyals2015show,karpathy2015deep,johnson2016densecap}.
Most of this work
generates
captions in English, but since the image captioning models do not
require linguistic knowledge, this is an arbitrary choice
based on the easy availability of training data in English. The same
captioning models are applicable for non-English languages as long as
sufficiently large training datasets are available
\cite{Elliott2015,Miyazaki2016}.

Of course, real image captioning applications will require support for
multiple languages.  It is possible to build a multilingual captioning
system by training a separate model for each individual
language, but this requires creating as many models as there are
supported languages.  A simpler approach would be to create a single
unified model that can generate captions in multiple languages.  This
could be particularly advantageous in resource-constrained
environments like mobile devices, where storing and evaluating
multiple large neural network models may be impractical. But to
what extent can a single model capture the (potentially) very
different grammars and styles of multiple languages?

\begin{figure}[t!]
  \centering
  \includegraphics[width =3in]{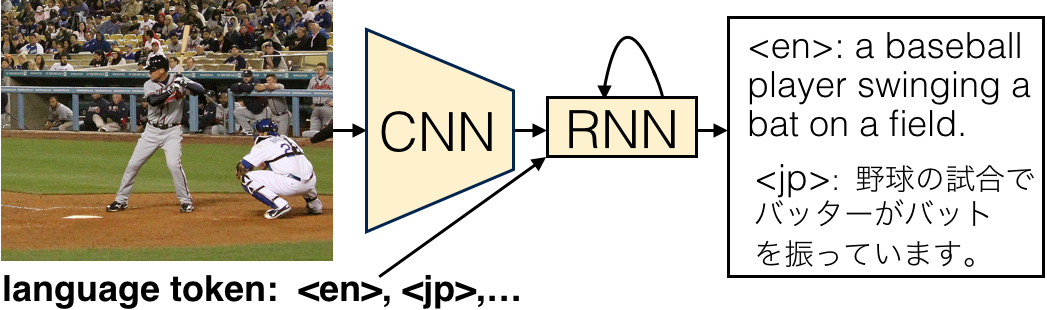}
\vspace{12pt}
 	\caption{In this work, we train a single model for multilingual captioning  using artificial tokens to switch languages.}
	\label{fig:overview}
\end{figure}

In this short paper, we propose training a unified caption
generator that can produce meaningful captions in multiple languages (Figure~\ref{fig:overview}).
Our approach is quite simple, but surprisingly effective: we inject an
artificial token at the beginning of each sentence to control the
language of the caption. During training, this special token informs
the network of the language of the ground-truth caption, while at test
time, it requests that the model produce a sentence in the specified
language.  We evaluated our approach using image captioning datasets
in English and Japanese. We chose Japanese because it reportedly has
the greatest linguistic distance from English compared to most other
popular languages, which means it is the most difficult language for a
native English speaker to learn~\cite{chiswick2005linguistic}. Our
experiments suggest that even for these two very distant languages, a
single neural model can produce meaningful multi-lingual image descriptions.

\section{Related Work}
The latest image captioning systems use multimodal neural networks,
inspired from sequence to sequence modeling in machine
translation~\cite{sutskever2014sequence}. Images are fed into a
Convolutional Neural Network (CNN) to extract visual features and
then converted to word sequences using a Recurrent Neural Network
(RNN) that has been trained on image-sentence ground truth
pairs~\cite{vinyals2015show,karpathy2015deep,johnson2016densecap}.

Most of this work on image captioning has considered a single target
language (English), although various other uses of multi-language
captioning data have been studied.
For instance, Elliott et al.~\cite{Elliott2015} and Miyazaki et al.~\cite{Miyazaki2016}
consider the problem of image captioning for one language when a caption
in another language is available.
Other work has considered the related but distinct problem of 
improving machine translation by using images that have been captioned
in multiple languages~\cite{Specia2016,Hitschler2016}. 
In contrast, our work assumes that we have training images that have been
captioned in multiple languages (i.e., each image has captions in multiple images,
and/or some images are captioned in one language and others are captioned in another),
and we wish to apply a single, unified model to produce captions in multiple languages on new, unseen
test images.

More generally, multilingual machine translation is an active
area of research.  The performance of machine translation can be
improved when training data in more than two languages is
available~\cite{Zoph2016MultiSourceNT,firat2016multi}, but the models
become more complex as the number of languages increases, because they
use separate RNNs for each language. The closest related work
to ours is Google's multilingual translation
system~\cite{johnson2016google} that uses artificial tokens to control
the languages. We apply a similar idea here for image caption generation.

\section{Model}
Our model uses a CNN to extract image features and then an RNN to
generate captions in a manner very similar to previous
work~\cite{vinyals2015show,karpathy2015deep}. Formally, we minimize
the negative log likelihood of the caption given an image,
\begin{equation}
 \min - \sum_{(I,S)} \sum_{t=0}^{|S|} \log p(S_t | I, S_0, \ldots, S_{t-1} ) , \label{loss}
\end{equation}
where each $(I,S)$ pair corresponds to an image $I$ and its caption
$S$, and $S_t$ is the $t$-th word of $S$. We use a
combination of CNN and RNN to model $p_t = p(S_t | I, S_0, \ldots,
S_{t-1})$ in the following manner:
\begin{eqnarray}
x_{-1} &=& \textrm{CNN}(I)\\
x_t &=& W_e S_t, \quad t\in\{0\ldots N-1\}\quad \\
p_{t+1} &=& \textrm{RNN}(x_t), \quad t\in\{0\ldots N-1\}\quad
\end{eqnarray}
where $W_e$ is a word embedding matrix, and $S_t$ is represented as a
one-hot vector. We use a special token assigned to $S_0$ to denote the start of the
sentence and the captioning language. For example, a monolingual captioning
model (a baseline) uses \texttt{<sos>} to indicate the starting of the
sentence, and the multilingual captioning model uses \texttt{<en>} or
\texttt{<jp>} to indicate English or Japanese, respectively. When
generating a caption, we find the sequence $S$ that (approximately) satisfies
equation (\ref{loss}) using a beam search.

\section{Experiments}
In order to investigate if it is possible to train a unified
captioning model across multiple languages, we experiment with
English and Japanese, which are the most
distant language pair among major languages~\cite{chiswick2005linguistic} and thus
should be particularly challenging. We evaluate the
quality of captions in English and Japanese under various models, including baseline
models trained only on individual languages, and the unified model trained with both.

\paragraph{Dataset.} We used the YJ Captions 26k Dataset~\cite{Miyazaki2016}, which is  based on a subset of MSCOCO~\cite{lin2014microsoft}. 
It has 26,500 images with Japanese captions. We also used the English
captions from the corresponding images in the original MSCOCO
dataset. We divided the dataset into 22,500 training, 2,000
validation, and 2,000 test images.

\paragraph{Implementation Details.} We segmented English sentences into tokens
based on white space, and segmented Japanese using
TinySegmenter\footnote{\url{https://github.com/SamuraiT/tinysegmenter}}. We
implemented a neural captioning model using the Chainer
framework~\cite{tokui2015chainer}, consulting the publicly-available
NeuralTalk implementation\footnote{\url{https://github.com/karpathy/neuraltalk2}}
for reference~\cite{karpathy2015deep}. For our CNN, we used
ResNet50~\cite{he2016} pre-trained on ImageNet, 
and an LSTM~\cite{hochreiter1997long} with
512 hidden units as our RNN. For training, we used stochastic gradient descent with the
Adam algorithm~\cite{kingma2014adam} with its default hyperparameters and
a batch size of 128. We trained 40 epochs, chose the model at
the epoch with the highest CIDEr validation score, and report
the performance on the test dataset. We used a beam size of 5 when
generating captions. We made the implementation publicly available\footnote{\url{http://vision.soic.indiana.edu/multi-caption/}}.

\begin{table}[!t]
\centering
\centering
\begin{tabular}{p{10mm}p{7mm}p{7mm}p{7mm}p{7mm}p{7mm}p{7mm}}
Train & Test & Bleu1 & Bleu2 & Bleu3 & Bleu4 & CIDEr \\
\hline
En & En & 0.632 & 0.443 & 0.298 & 0.119 & 0.651 \\
En+Jp & En & 0.558 & 0.410 & 0.275 & 0.184 & 0.593 \\
\hline
Jp & Jp & 0.698 & 0.553 & 0.410 & 0.303 & 0.631 \\
En+Jp  & Jp & 0.693 & 0.534 & 0.414 & 0.310 & 0.594 \\
\hline
\end{tabular}
\vspace{8pt}
 \caption{Evaluation of captioning results.}
\label{tbl:results}
\end{table}

\paragraph{Results}
Table \ref{tbl:results} reports the results of our various models
using Bleu scores~\cite{papineni2002bleu}, which are designed to evaluate machine
translation, and the CIDEr score \cite{vedantam2015cider}, which is
designed to evaluate image captioning. Bleu uses the overlap of $n$-grams between the predicted sentence and ground truth sentences; we follow typical practice and report for four different values of $n$. The results show that Bleu scores decrease slightly from the single language models to the dual-language models, but
are very close (and in fact the dual model performs better than the single language models for 3 of the 8 Bleu variants). Under CIDEr,
the multi-language models again perform slightly worse
(e.g. 0.651
vs 0.593 for English and 0.631 versus 0.594 in Japanese).
However, the single unified models effectively use half as much memory
and require half as much time to generate captions.
This is a promising result: despite being trained on very
distant languages, the scores indicate that the model can still produce meaningful, linguistically correct captions, with 
a nearly trivial change to the training and caption generation algorithms. 

\begin{figure*}[p]
  \centering
  \includegraphics[width =160mm]{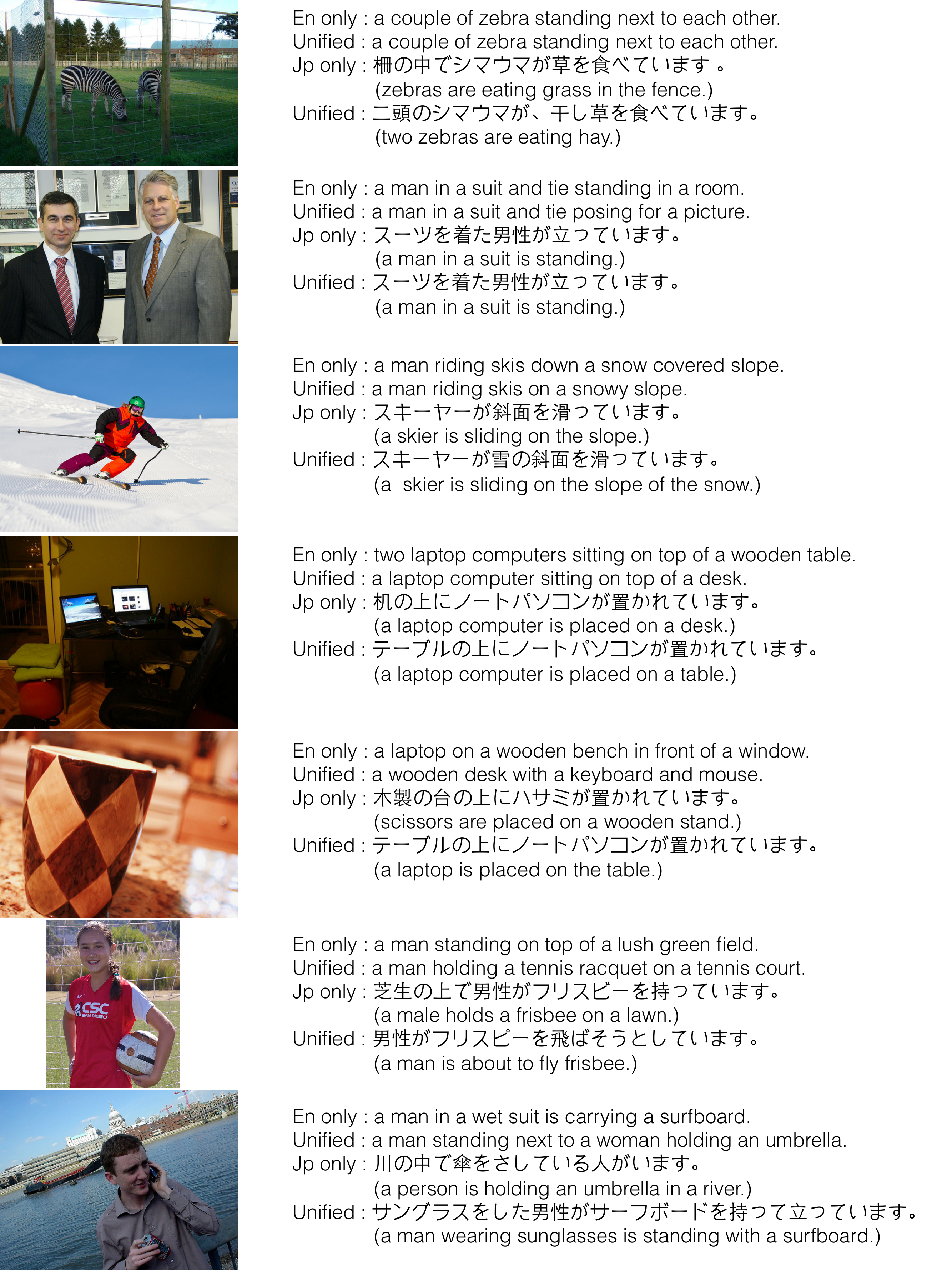}
 	\caption{Randomly selected samples of automatically-generated image captions. 
 \textit{En only} captions are from 
the model trained on English, \textit{Jp
          only} are from the model trained on Japanese, and \textit{Unified} are from the single model trained with both.}
	\label{fig:results}
\end{figure*}

Some randomly selected captions are shown in
Figure \ref{fig:results}. We qualitatively observe that the single and dual-language models tend to make similar types of mistakes. For example, all captions for the bottom picture incorrectly mention water-related concepts such as surfboards and umbrellas, presumably due to the prominent water in the scene. 

\section{Conclusion}
We introduced a simple but effective approach that uses artificial tokens
to control the language for a multilingual image captioning system. Our
approach is simpler and more compute- and memory-efficient than having to build a separate model for each language, by allowing the system to share a single
model across multiple languages.  Our preliminary experiments suggest
that a neural captioning architecture can learn a
single model even for English and Japanese, two languages whose linguistic distance is especially
high, with minimal decrease in accuracy compared to individual models. 
In future work, we plan to investigate further refinements that would
remove these accuracy gap altogether, as well as to evaluate on multiple languages
beyond just English and Japanese.

\section*{Acknowledgment}
Satoshi Tsutsui is supported by the Yoshida Scholarship Foundation in Japan. This work was supported in part by the National Science Foundation (CAREER IIS-1253549) and the
IU Office of the Vice Provost for Research and the School of Informatics and Computing through the Emerging Areas of Research Funding Program,
and used the Romeo FutureSystems Deep Learning facility, supported by Indiana University and NSF RaPyDLI grant 1439007. 

{\small
\bibliographystyle{ieee}
\bibliography{references}
}

\end{document}